\documentclass[10pt,twocolumn,letterpaper]{article}

\usepackage[pagenumbers]{cvpr}              

\usepackage{graphicx}
\usepackage{amsmath}
\usepackage{amssymb}
\usepackage{booktabs}
\usepackage{enumitem}
\usepackage{bm}
\usepackage[accsupp]{axessibility}

\usepackage[pagebackref,breaklinks,colorlinks]{hyperref}

\usepackage[capitalize]{cleveref}
\crefname{section}{Sec.}{Secs.}
\Crefname{section}{Section}{Sections}
\Crefname{table}{Table}{Tables}
\crefname{table}{Tab.}{Tabs.}

\usepackage[many]{tcolorbox}

\def\confName{CVPR}
\def\confYear{2024}

\begin{document}


\title{A Survey of Video Datasets for Grounded Event Understanding}

\author{Kate Sanders\\
Johns Hopkins University\\
{\tt\small ksande25@jhu.edu}
\and
Benjamin Van Durme\\
Johns Hopkins University\\
{\tt\small vandurme@jhu.edu}
}
\maketitle

\begin{abstract}
   While existing video benchmarks largely consider specialized downstream tasks like retrieval or question-answering (QA), contemporary multimodal AI systems must be capable of well-rounded common-sense reasoning akin to human visual understanding. A critical component of human temporal-visual perception is our ability to identify and cognitively model ``things happening", or events. Historically, video benchmark tasks have implicitly tested for this ability (e.g., video captioning, in which models describe visual events with natural language), but they do not consider video event understanding as a task in itself. Recent work has begun to explore video analogues to textual event extraction but consists of competing task definitions and datasets limited to highly specific event types. Therefore, while there is a rich domain of event-centric video research spanning the past 10+ years, it is unclear how video event understanding should be framed and what resources we have to study it. In this paper, we survey 105 video datasets that require event understanding capability, consider how they contribute to the study of robust event understanding in video, and assess proposed video event extraction tasks in the context of this body of research. We propose suggestions informed by this survey for dataset curation and task framing, with an emphasis on the uniquely temporal nature of video events and ambiguity in visual content.
\end{abstract}


\section{Introduction}

\begin{figure}[ht!]
  \includegraphics[width=.49\textwidth]{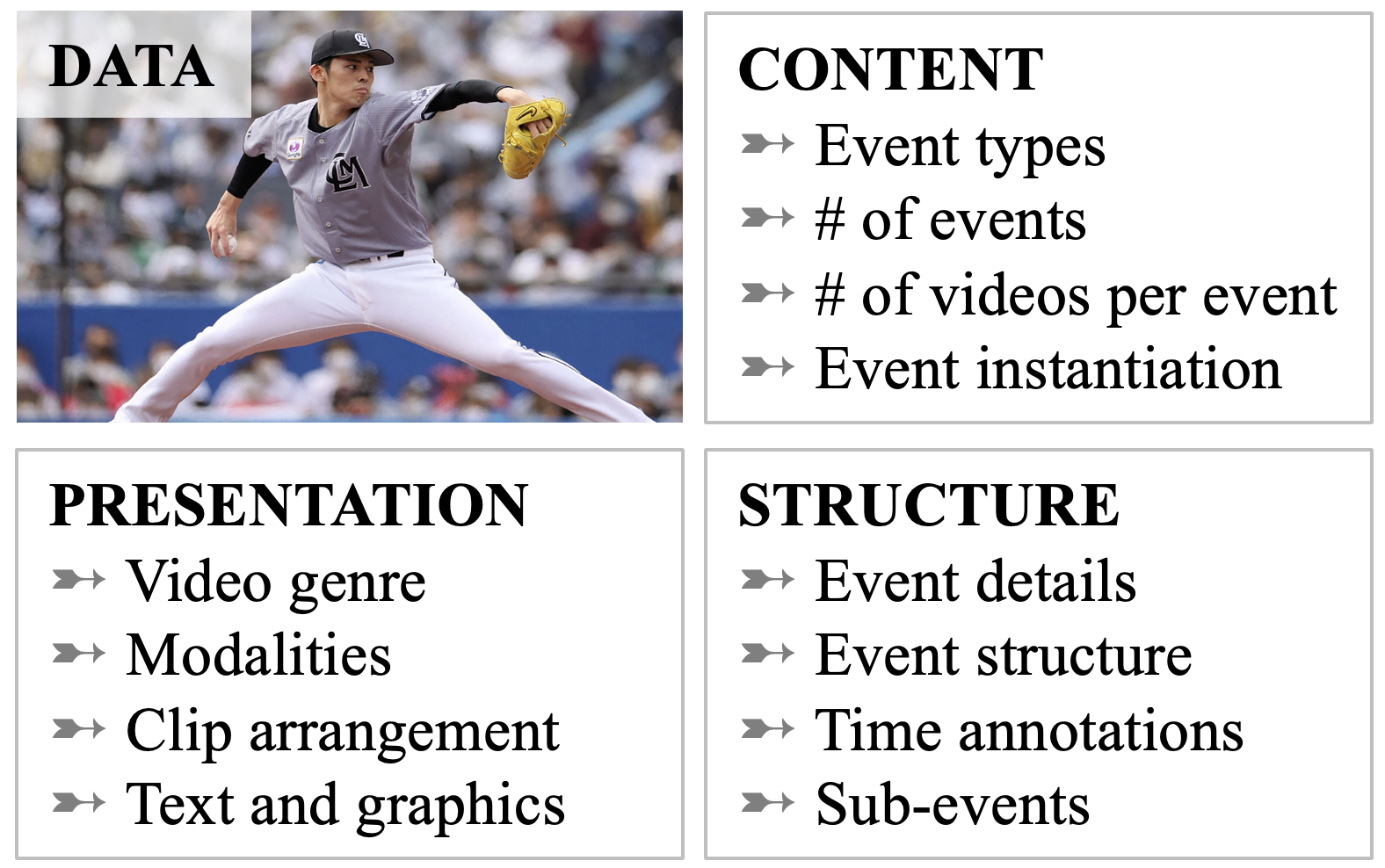}
  \caption{Illustration of an example video dataset paired with the associated topics we cover in our survey, detailed in Section 2. These topics are designed to help us answer the question of what resources we have for robust video event understanding, targeting the \textit{content}, \textit{presentation}, and \textit{structure} of video events.}
  \label{fig1}
\end{figure}

Until recently, video understanding has focused on specialized models capable of highly specific downstream tasks \cite{vahdani2021deep, 9350580, 10.1016/j.engappai.2020.103557, kong2022human, 9594911}. As general-purpose AI assistants like Gemini have begun to accept video inputs \cite{reid2024gemini}, it is no longer sufficient to rely on suites of video models that are tailored to perform well on a small set of video benchmarks. The next generation of video models will need to facilitate robust, generalizable, and human-like visual reasoning over unconstrained video inputs and ground them to natural language text. These systems, then, necessarily require an ability to interpret and synthesize low-level visual features into complex semantic ideas that humans care about. A key component of this is the fundamental task of identifying and classifying ``things that are happening", notions often modeled in natural language processing as ``events" \cite{maienborn2011event}. The purpose of this survey is to assess the breadth of video datasets and tasks introduced in the last decade to answer the question: \textit{Given the goal of developing vision-language models that can interpret events depicted in video as humans can, what resources do we have for working towards this goal, and are they sufficient?}

There are a variety of datasets and approaches for event extraction in text documents \cite{liu2021overview, 8918013, 9927311, 9627684}. Some researchers have leveraged the distinct relationship between visual content and textual event semantics to explore visual event semantics for single images, ranging from k-means classification tasks to semantic role labeling and grounding \cite{gupta2015visual, yatskar2016situation, pratt2020grounded}. Comparatively, the work in video event understanding is understudied and less structured, despite the richness introduced to the domain through the temporality of video. While many video benchmarks involve identifying some form of events, the domain, semantic complexity, and representation of events diverge substantially across benchmarks as the primary focus is typically on their downstream task. The few datasets that do introduce event extraction as the core task propose diverging task descriptions and limited data to facilitate their study \cite{chen2021joint, sadhu2021visual, ayyubi2022multimodal}. This makes it difficult to establish a notion of what resources the field has for the study of event understanding, and difficult to establish a notion of what an event should even be in the context of video data.

\begin{figure*}
  \includegraphics[width=\textwidth]{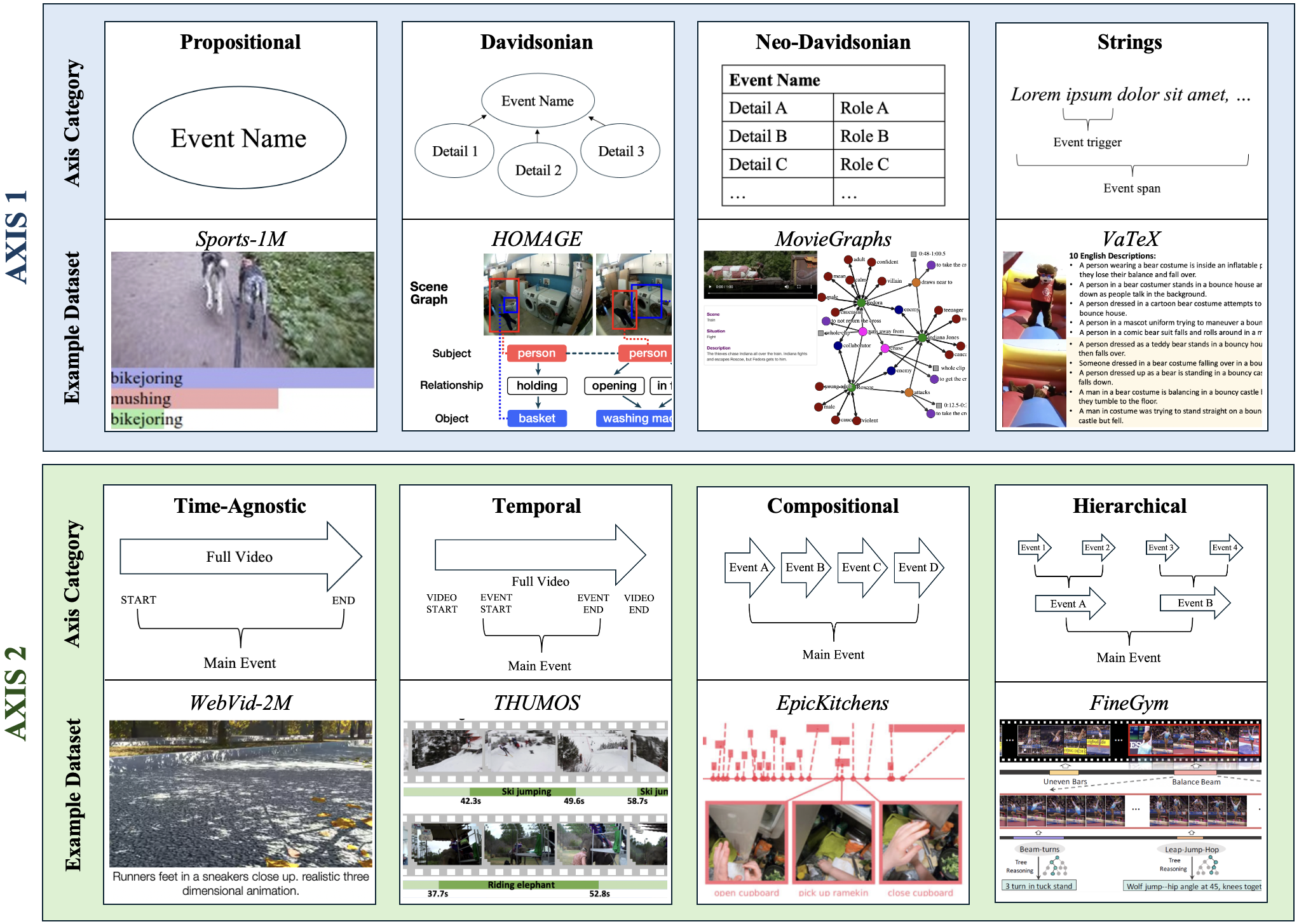}
  \caption{Illustration of the two axes along which we classify event structures presented in video datasets. Alongside each category name, we include a graphic depicting the general idea of the method, a dataset that falls into this structure category, and an image of the dataset presentation to show how these methods exist in real data (dataset images are taken from the original publication). The top row shows how details are organized, and the bottom row shows how time is incorporated into the structure.}
  \label{fig2}
\end{figure*}

These uncertainties motivate our survey. We construct a framework to analyze video datasets with a focus on (1) what events are presented, (2) how they are presented, and (3) how they are interpreted, pictured in Figure \ref{fig1}. We consider a set of 105 video datasets and organize them into categories based on their intended task and construction approach. We analyze these datasets using our framework and provide a summary of our primary findings, and provide a dataset analysis to the community that will allow users to identify datasets that meet their needs.\footnote{Data released at \url{https://github.com/katesanders9/grounded-events}} We then introduce a taxonomy of video event structures (Figure \ref{fig2}) to study the representational differences among events in different video datasets. We draw from this analysis to assess the relationships between existing video event extraction benchmarks and the surveyed datasets. Finally, we make suggestions for future dataset construction and video event understanding formulation, focusing on the temporal structure of video events, the relationship between formal event structures and natural language descriptions, and uncertainty in video understanding.

\section{Survey Methodology}
\label{s2}

To guide our analysis of video understanding resources, we establish three primary dataset considerations: What events are presented (content), how they are presented (presentation), and how they are interpreted (structure).

\subsection{Dataset Content}

As we are concerned with studying events through video, we must assess what events are in existing datasets. We should consider not only the types of events recorded, but how the recorded events occur, because naturally occurring events often have notable visual differences from staged content for TV shows, or events recreated for a benchmark. For each dataset, we ask:

\begin{itemize}[leftmargin=*]
\setlength\itemsep{0em}
\item \textbf{What is the domain of events depicted?} (Sports, daily activities, social interactions, \ldots)
\item \textbf{How many events are depicted?} (100 activity instances, 4K videos, 70 event classes, \ldots)
\item \textbf{Are there multiple videos included per event?} (Yes/No)
\item \textbf{How were the events instantiated?} (Natural or staged)
\end{itemize}

\subsection{Dataset Presentation}

Video content should not only be assessed for the semantic content it depicts but how it is presented through the audio-visual medium. This assessment largely pertains to choices surrounding the filming and editing of videos, such as filming purpose, video quality, and additional content added in post. We ask four general questions to characterize the presentation of visual events:

\begin{itemize}[leftmargin=*]
\setlength\itemsep{0em}
\item \textbf{What is the purpose of the videos?} (Home videos, news broadcasts, dataset content, \ldots)
\item \textbf{What modalities are recorded?} (Video only, video + audio, video + depth info, \ldots)
\item \textbf{Are the videos compilations?} (Single video stream or concatenated clips)
\item \textbf{Are text or graphics added in post?} (Yes/No)
\end{itemize}

\subsection{Dataset Structure}
Event-centric videos can be explained through many distinct conceptions of events. It is important to identify what semantic event structures existing benchmarks implicitly promote, as they dictate what we train vision models to care about and how to structure information. We find that existing structures can be understood through two distinct axes: how event details are selected and organized, and how the temporal nature of videos are incorporated. A sample of video event structures identified in our survey and their relationships to these axes are illustrated in Figure \ref{fig2}. To analyze dataset event structures, we ask the questions:

\begin{itemize}[leftmargin=*]
\setlength\itemsep{0em}
\item \textbf{What event details are included?} (Actions, subjects and actions, event templates, \ldots)
\item \textbf{How are the details organized?} (Event structures or natural language strings)
\item \textbf{How is the timing of events structured?} (No annotations, start and end points, or multiple events with temporal relationships)
\item \textbf{Are hierarchical events considered?} (Yes/No)
\end{itemize}
\section{Dataset Collection}
Hundreds of video datasets have been released in the last decade.  We limit the scope of considered datasets to those whose task description involves identifying or reasoning over at least one action, omitting datasets with applications strictly concerning ``low-level" features like shapes, textures, composition, individual objects, trajectories, and attributes. This includes tasks such as object tracking, texture detection, and segmentation.

Many of the answers to questions detailed in Section \ref{s2} are highly correlated with the dataset construction method, which is often dictated by the benchmark task. So, in our survey we organize datasets by general application category, and when applicable, unique construction methodology. We identify five primary application categories: Basic action recognition, classification, and localization; hierarchical action recognition, classification, and localization; scene parsing and entity relation detection; retrieval and captioning; and question-answering.

We find that videos surveyed typically belong to one of the following four production formats: Uncurated non-professional footage (YouTube uploads, vlogs, etc.); curated non-professional footage (e.g., scripted scenarios acted out by humans recruited for dataset construction); professional news footage (broadcasts covering real-world events); and professional narrative content (TV shows and movies). The latter two formats produce unique, format-specific artifacts that influence the content of the dataset compared to unprofessional footage. In a similar vein, we find that multilingual video datasets typically follow unique and specific construction methods as well. So, we add three additional categories: movie and TV show footage; news broadcasts; and multilingual content.

Finally, a small number of datasets do not fall under any of the discussed categories, and so we classify these as miscellaneous datasets as a final ninth category.

We search through research paper and dataset literature databases to identify a set of 105 datasets that have either been released or frequently used in the last ten years. We include a list of these datasets in Table \ref{tab1}.

\begin{table*}
    \centering
    \begin{tabular}{lp{2in}p{3in}}
        \textbf{Category} & \textbf{Description} & \textbf{Datasets} \\
        \hline
        Action Recognition & \small{Typically short videos of humans performing singular actions. Some datasets consider a complex definition of ``actions" and include things like group activities.} & \small{UCF-101\cite{soomro2012ucf101}, Kinetics\cite{carreira2019short}, MomentsInTime\cite{monfort2019moments}, Multi-MomentsInTime\cite{monfort2021multi}, Sports-1M\cite{karpathy2014large}, THUMOS\cite{idrees2017thumos}, ActivityNet\cite{caba2015activitynet}, Charades\cite{sigurdsson2016hollywood}, Charades-Ego\cite{sigurdsson2018actor}, MMAct\cite{kong2019mmact}, VideoLT\cite{zhang2021videolt}, YouTube-8M\cite{abu2016youtube}, HVU\cite{diba2020large}, VLOG\cite{fouhey2018lifestyle}, DiDeMo\cite{anne2017localizing}, HLVU\cite{curtis2020hlvu}, Volleyball\cite{ibrahim2016hierarchical}, NBA\cite{yan2020social}, CCV\cite{jiang2011consumer}, FCVid\cite{jiang2015fcvid}, ShanghaiTech Campus\cite{luo2017revisit}, Animal Kingdom\cite{ng2022animal}, Assembly101\cite{sener2022assembly101}, Something-Something\cite{goyal2017something}, HMDB51\cite{6126543}, AVA Actions\cite{gu2018ava}, MTL-AQA\cite{parmar2019and}, HowTo100M\cite{miech2019howto100m}, Multi-HowTo100M\cite{huang2021multilingual}} \\
        \hline
        
        Hierarchical Recognition & \small{Annotations structure events as ``building blocks" through which complex, abstract events can be structured.} & \small{Homage \cite{rai2021home}, LEMMA \cite{jia2020lemma}, FineGym \cite{shao2020finegym}, MultiTHUMOS \cite{yeung2018every}, COIN \cite{tang2019coin}, MPII Cooking2 \cite{rohrbach2016recognizing}, Epic Kitchens \cite{damen2018scaling}, YouCook2 \cite{zhou2018towards}, Multi-YouCook2 \cite{rouditchenko2023c2kd}, MUSES \cite{liu2021multi}, Breakfast \cite{kuehne2014language}, Diving48 \cite{li2018resound}, NewsNet \cite{wu2023newsnet}} \\
        \hline
        
        Scene Graphs & \small{Annotations place a strong emphasis on relationships between many entities and/or actions in videos.} & \small{MOMA \cite{luo2021moma}, VidSitu \cite{sadhu2021visual}, VM2E2 \cite{chen2021joint}, TVEE \cite{wang2023cross}, MovieNet \cite{huang2020movienet}, MovieGraphs \cite{vicol2018moviegraphs}}, AIDA Scenario 1 \cite{aida} \\
        \hline
        
        Retrieval/Captioning & \small{Videos of variable complexity, content, and genre, generally paired with unique natural language descriptions.}  & \small{VaTeX \cite{wang2020vatex}, WebVid-2M \cite{zhou2019grounded}, VTW \cite{zeng2016leveraging}, TGIF \cite{li2016tgif}, V3C \cite{rossetto2018v3c}, MSR-VTT \cite{xu2016msr}, VideoStory \cite{habibian2014videostory}, ActivityNet-Captions \cite{caba2015activitynet}, ActivityNet-Entities \cite{zhou2019grounded}, TVR \cite{lei2020tvr}, TVC/TVR \cite{lei2020tvr}, TRECVID \cite{smeaton2006evaluation}, EventNet \cite{ye2015eventnet}, MultiVENT \cite{sanders2023multivent}, Marine Video Kit \cite{truong2023marine}, InternVid \cite{wang2023internvid}, MSVD \cite{chen2011collecting}, FIVR-200K \cite{kordopatis2019fivr}, VITT \cite{huang2020multimodal}, Spoken MomentsInTime \cite{monfort2021spoken}, VideoCC3M \cite{nagrani2022learning}, MultiVENT \cite{sanders2023multivent}, TextVR \cite{wu2023large}, MPII \cite{rohrbach2015dataset}, MAD \cite{DBLP:journals/corr/abs-2112-00431}, CondensedMovies \cite{bain2020condensed}, HowTo100M \cite{miech2019howto100m}, Multi-HowTo100M \cite{huang2021multilingual}, mTVR \cite{lei2021mtvr}, VATEX \cite{wang2020vatex}, How2 \cite{9747320}, RUDDER \cite{a2021rudder}, ChinaOpen \cite{Chen_2023}} \\
        \hline
        
        Q\&A & \small{(Typically) semantically complex videos with paired natural language question-answer pairs. Sometimes the questions are multiple choice.} & \small{MSR-VTT-QA \cite{xu2017video}, HowToVQA69M \cite{miech2019howto100m}, TVQA \cite{lei2019tvqa}, MovieQA \cite{tapaswi2016movieqa}, NEXT-QA \cite{xiao2021nextqanext}, NewsVideoQA \cite{jahagirdar2023watching}, SocialIQ \cite{zadeh2019social}, EgoSchema \cite{mangalam2024egoschema}, Perception Test \cite{patraucean2024perception}, DramaQA \cite{choi2021dramaqa}, AGQA \cite{grunde2021agqa}, STAR \cite{wu2021star}, Ego4D \cite{grauman2022ego4d}} \\
        \hline
        
        Movie/TV Shows & \small{Datasets of video clips taken from content produced to convey some sort of narrative to humans. Annotations vary substantially between datasets.} & \small{HLVU \cite{curtis2020hlvu}, MPII \cite{rohrbach2015dataset}, MovieNet \cite{huang2020movienet}, MAD \cite{DBLP:journals/corr/abs-2112-00431}, MovieGraphs \cite{vicol2018moviegraphs}, CondensedMovies \cite{bain2020condensed}, TVR \cite{lei2020tvr}, TVC \cite{lei2020tvr}, TVQA \cite{lei2019tvqa}, TVQA+ \cite{lei2020tvqa}, MovieQA \cite{tapaswi2016movieqa}} \\
        \hline
        
        News & \small{Video clips taken from content produced specifically to convey news and current events to humans. Often OCR-heavy.} & \small{TextVR \cite{wu2023large}, NewsVideoQA \cite{jahagirdar2023watching}, NewsNet \cite{wu2023newsnet}, VM2E2 \cite{chen2021joint}, TVEE \cite{wang2023cross}, AcTiV \cite{7333911}, MultiVENT \cite{sanders2023multivent}} \\
        \hline
        
        Multilingual & \small{Datasets building on existing datasets with the explicit goal of multilinguality, often by adding new multilingual annotations.} & \small{Multi-YouCook2 \cite{rouditchenko2023c2kd}, Multi-HowTo100M \cite{huang2021multilingual}, mTVR \cite{lei2021mtvr}} \\
        \hline
        
        Misc & \small{Other datasets of note that do not fall into a traditional task category. Many of these datasets facilitate multiple tasks and so their construction is unique.} & \small{Spoken Moments \cite{monfort2021spoken}, HACS \cite{zhao2019hacs}, TVSum \cite{song2015tvsum}, TEMPO \cite{hendricks2018localizing}, How-2 \cite{sharma2022end}, QuerYD \cite{oncescu2021queryd}, VIOLIN \cite{liu2020violin}, CrossTask \cite{zhukov2019crosstask}, HD-VILA-100M \cite{xue2022advancing}, VALUE \cite{li2021value}, VLEP \cite{lei2020likely}, MultiHiEve \cite{ayyubi2022multimodal}, 3MASSIV \cite{gupta20223massiv}, VEATIC \cite{ren2024veatic}, EmoMV \cite{thao2023emomv}, VideoInstruct \cite{maaz2023video}} \\
        \hline
    \end{tabular}
    \caption{Breakdown of surveyed datasets by task type, with brief summaries describing the dataset type.}
    \label{tab1}
\end{table*}
\section{Video Analysis}
Below are brief summaries of general trends present in the datasets we survey, organized based on our ``content" and ``presentation" considerations. 

\subsection{Results}

\noindent\textbf{Action Recognition}\hspace{2mm}
Traditional action recognition benchmarks often largely consider simple events, building on work in pose recognition and action recognition in the image domain. However, a number of datasets have extended the concept of action recognition to more complex “actions” including longer, abstract events like “group activity” recognition. Many of these datasets are limited in scope, considering a small domain of event types, like ``diving"~\cite{li2018resound}, but some include a wider domain with the aim of covering a broad range of daily human experiences.\\

\noindent\textbf{Hierarchical Action Recognition}\hspace{2mm}
Some action recognition datasets go beyond general event recognition by decomposing actions into hierarchies of sub-actions. Cooking video datasets in particular lend themselves well to this domain. Like the standard action recognition datasets, and especially due to the level of annotation detail necessary, most of these datasets (e.g., cooking datasets) do not consider a wide range of content. MultiTHUMOS \cite{yeung2018every} is notable in its significant range of event types.\\

\noindent\textbf{Scene Graphs}\hspace{2mm}
Scene graph datasets often involve visual relation extraction and include annotations for objects’ relations to each other and their locations within the video - this tends to push the dataset focus toward more complex content. These datasets often include many event details and include detailed temporal/spatial annotations.\\

\noindent\textbf{Retrieval and Captioning}\hspace{2mm}
A number of video retrieval datasets focus on short and simple events that can be depicted in a few seconds, but others include long and sophisticated content. The length of the paired captions generally dictate the complexity of the events, for example when video content must be able to be described in a single sentence, but this is not always the case. Furthermore, the range of event types is often quite large.\\

\noindent\textbf{Question-Answering}\hspace{2mm}
Question-answering datasets are often built by re-annotating existing video datasets, often by leveraging natural language captions or narration. QA datasets lend well to narratively and socially complex video content often produced for TV, whereas other QA datasets follow more traditional formats that focus less on high-level reasoning. This latter set of QA datasets may consider a wider range of video genres than those focusing on TV content, aligning more with the range seen in captioning datasets described above.\\

\noindent\textbf{Movie/TV Datasets}\hspace{2mm}
Building on the overview of QA datasets, a number of other datasets depicting movies and TV shows have been introduced. Naturally, high-complexity narrative videos uniquely facilitate complex, narrative-driven reasoning work, and a high proportion of these datasets include temporal and spatial annotations.\\

\noindent\textbf{Multilingual Content}\hspace{2mm}
Researchers typically build multilingual datasets by either directly translating existing datasets or, sometimes, collecting new, organic multilingual data. Datasets collected from organic multilingual content are often limited in genre, with a focus on studying the multilinguality of one specific domain.\\

\noindent\textbf{News Videos}\hspace{2mm}
It is difficult to generate text documents for video content, but news videos allow for easy ``real world" text document mapping. Most of these datasets are limited in presentation format but can include rich on-screen textual data.\\

\subsection{Discussion}
\label{s42}
The large majority of surveyed datasets consider semantically simpler events than those typically found in text-based event extraction datasets \cite{grishman1996message, jain2020scirex, vashishtha2023famus}. The datasets that do include complex event types tend to have a limited range of event genres, and the videos often come from highly edited or staged content like TV shows. While teaching models to understand low-level events is a necessary step for understanding more complex events, as vision models continue to improve it will become increasingly necessary to have datasets that present a wide range of complex event types.

On a similar note, it is difficult to collect data from a wide range of presentation formats using traditional video dataset construction approaches (i.e. not using synthetic videos generated by models), and in some tasks, it is unnecessary. This is reflected in the datasets surveyed: Very few consider multiple filming or editing methods, and many do not include multiple filming locations, setups, or participants. However, as video-language models for daily task assistance grow commonplace, it will grow increasingly important for models to be familiar with a wide range of video presentation methods to mitigate geographic, linguistic, and cultural bias. Therefore, we argue that training models on datasets constructed through multiple presentation methods is critical, which necessitates being conscientious of such considerations when constructing datasets.

Surveyed datasets with a high level of semantic complexity sometimes require event understanding beyond identifying the most likely events present in a video. Social reasoning datasets may require high-level probabilistic inference, and multiple choice QA datasets may require comparing two likely answers to identify which is \textit{better}. This distinction between basic event understanding and high-level inference over possible events is important when considering the type of event understanding necessary for a task.
\section{Structure Analysis}

To ascertain how vision models are expected to understand events, we consider how events are structured in each benchmark. We find that semantic annotations make it easy to isolate the concrete event information necessary to perform a dataset task, agnostic to the annotation format through which the information is presented. Below, we outline past work in studying event structures and connect this work to video events. The axes are illustrated in Figure \ref{fig2} alongside example datasets.

\subsection{Events and Language}
Davidson is credited \cite{rothstein2001events, maienborn2011event} as an early influential proponent for considering events as spatiotemporal things, proposing a linguistic framework through which they could be modeled formally \cite{davidson1967logical}, alongside Reichenbach who proposed related notions involving logical forms for events \cite{reichenbach1947elements}. Since then, linguists have built on these ideas, introducing model extensions \cite{parsons1990events, higginbotham1983logic, schein1993plurals, 10.5555/374875.374896} and adapting the core ideas to tasks in natural language processing \cite{hogenboom2011overview}. A primary application of event semantics is the extraction of events within text documents. Much of this work is based on the idea of defining events through frames (templates) \cite{baker1998berkeley}. A variety of datasets for this task have been introduced \cite{chinchor1998overview, li2020duee, gantt2024multimuc} and robust methods have been proposed \cite{liu2021overview, gao2023exploring, chen2022iterative}.

Some researchers have leveraged the distinct relationship between visual content and textual event semantics to explore visual event semantics. Initially, visual event recognition was limited to k-way classification tasks \cite{soomro2012ucf101, gupta2009observing, everingham2015pascal}, but Gupta et al. \cite{gupta2015visual} expanded the definition of a visual event through the introduction of "visual semantic role labeling", or the task of mapping images to a limited taxonomy of scenarios following a subject-object-action template. A ``grounded" version of this task was proposed by Yang et al. \cite{yang2016grounded}. These ideas were built on by Yatskar et al. \cite{yatskar2016situation} and Pratt et al. \cite{pratt2020grounded} who presented a more complex task of mapping images to template-based scenarios, but with a much wider range of templates following the FrameNet ontology \cite{baker1998berkeley}.

In the following sections, we outline the similarities and differences between these event structures in text and images and the event structures necessary for explaining video event benchmarks. We find that there are two primary axes along which video event interpretations diverge across datasets, which we illustrate through a set of formal semantic annotations. 

\subsection{Semantic Structure}
The first axis considers how the details of the event are organized. We can understand this axis through a comparison to event structuring in formal semantics. Below, we characterize differences among surveyed detail organization approaches. For each category, we consider the video depicted in Figure \ref{fig1} (a video of a pitcher throwing a ball at a baseball game) as a running example and outline how this video might be structured through each primary framework.\\

\noindent\textbf{Propositional Structures}\hspace{2mm}
The simplest semantic structures only consider the verb: Events are structured as propositions with no defined details. The video of the baseball game, therefore, would simply be structured as
\begin{equation}
Throwing
\end{equation}
This modeling framework often applies to datasets depicting very simple events, such as action recognition datasets in which each video is mapped to a short label.\\

\noindent\textbf{Davidsonian Structures}\hspace{2mm}
More complex modeling methods include a set of details for each event in the dataset. In some benchmarks, the same set of details is considered for each event, regardless of semantic distinctions between them (generally a proto-agent, proto-patient, and predicate). The example video, then, could be structured as
\begin{equation}
\exists\, e,x,y. \;\, \textit{Throwing}\,(\textit{pitcher}\,(x)\,,\textit{baseball}\,(y)\,,\,e)
\end{equation}
More complex action recognition and hierarchical action recognition datasets often fall into this category, but the majority of mid-level complexity datasets require more complex structures to properly represent the events.\\

\noindent\textbf{Neo-Davidsonian Structures}\hspace{2mm}
A third class of structures similarly structures the details associated with events, but treats them as variable ``template fields" that may differ depending on event type. Each detail associated with an event is typically labeled with a ``template slot type", used to define the semantic relationship between itself and the encompassing event. Using this framework, the baseball pitch could be structured as
\begin{multline}
\exists\, e,x,y,z.\, [\,\textit{Throwing}\,(e)\, \wedge\; \textit{Agent}\,(\textit{pitcher}\,(x)\,, e)\, \wedge\\ \textit{Obj}\,(\textit{baseball}\,(y)\,, e)\, \wedge\;\textit{At}\,(\textit{third inning}\,(z)\,, e)\; ...]
\end{multline}
This structure covers many datasets with more complex annotations. The majority of mid-tier complexity event datasets fall into this category, as they can be structured fairly easily through this structure.\\

\noindent\textbf{String-based Structures}\hspace{2mm}
Finally, many datasets do not include formally organized structures of the depicted events. Instead, they are structured only using natural language descriptions, or strings. There generally is not an established structure that the strings abide by, and so there is typically no way to ensure which structuring approach is most appropriate for an entire dataset of labels. The example video could be represented as
\begin{equation}
\textit{The pitcher throws the ball during the third inning.}
\end{equation}
The majority of captioning, retrieval, and QA datasets fall into this category, as videos in these datasets typically are paired with long-form natural language descriptions (or in the case of QA, a two-part or multi-part natural language description split across questions and answers). The events in this category tend to be more complex than those covered by the other event structures.

\subsection{Semantic Structure: Discussion}
It should be noted that while these categories can help us categorize and understand visual event structures, they do not serve as complete explanations: Two event structures that both classify as Neo-Davidsonian structures may differ in their detail and adherence to traditional Davidsonian structuring methods, and the same applies to the temporal axis organizational framework. We argue that Davidsonian and Neo-Davidsonian event representations are superior to propositional structures, as they reduce ambiguity and underspecification. But, based on the surveyed datasets, the details of the structuring method are domain-dependent.

There is a notable divide between datasets in terms of whether they include formal event structures or natural language string descriptions. Few datasets include both. We argue that formal representations and fully string-based representations are both important for robust visual event understanding. Formal structures are useful for evaluation and organization, while natural language allows for better human alignment. 

\subsection{Temporal Structure}
The second axis concerns how time within individual videos influences the event structure, which diverges from most work in traditional language semantics. Again, we consider four main categories into which the surveyed event structures fall, and consider how our structures can be organized using this framework.\\

\noindent\textbf{Time-Agnostic Structures}\hspace{2mm}
Some benchmarks do not structure the temporal position at which events exist in a given video: One video simply corresponds to one event. This can roughly be structured as
\begin{equation}
\exists\, e. \;\, \textit{Event}\,(e)
\end{equation}
Some action recognition datasets are structured this way, and many captioning and retrieval datasets involve short enough videos that time annotations are unnecessary.\\

\noindent\textbf{Temporal Structures}\hspace{2mm}
On the other hand, many datasets' event structures include the time with respect to the full input video at which an event takes place:
\begin{equation}
\exists\, e,t.\;\, [\,\textit{Event}\,(e)\, \wedge\; \textit{At}\,(e,\,t_{1},t_{2})\,]
\end{equation}
Action localization datasets, a subset of action recognition datasets, generally use temporal event structures. Some other datasets also structure events this way, especially those using string-based methods. For these datasets, $t$ is sometimes loosely defined, but still defined nonetheless.\\

\noindent\textbf{Compositional Structures}\hspace{2mm}
Some benchmarks include multiple ``sub-events" within the same video. While these sub-events are sometimes referred to as full events within the task or dataset description, we argue that through this formulation, more semantically interesting events are implicitly modeled within videos and task outputs, warranting a distinction. The structuring of these events can be illustrated as
\begin{multline}
\exists\, e,t.\;\, [\,\textit{Event}\,(e_x)\, \wedge\; \textit{At}\,(e_x,\,t_{1(x)},t_{2(x)})\,\wedge\;\\ \textit{Event}\,(e_y)\, \wedge\; \textit{At}\,(e_y,\,t_{1(y)},t_{2(y)})\,]
\end{multline}
The majority of surveyed dataset structures are compositional in nature, especially those with natural language descriptions. However, string-based structures are difficult to classify into one temporal approach or the other due to the fact that there is often little structure incentivizing consistent temporal labeling across all videos in a dataset.\\

\noindent\textbf{Hierarchical Structures}\hspace{2mm}
Building on the ideas introduced by compositional structures, hierarchical structures also consider multiple ``sub-events" within a single video \cite{nguyen2023resin}. However, through this structuring paradigm, the sub-events are explicitly defined as sub-events, and multiple layers can be established. The compositional complexity of these structures differs among approaches. For its formulation, we use the predicate \textit{Comprises}(\textit{x},\textit{y},\textit{z}) where sub-events \textit{x} and \textit{y} exist within full event \textit{z} (this predicate can be further formalized through episodic logic \cite{i1993episodic} as $[x*z]\wedge [y*z]$):
\begin{multline}
\exists\, e,t.\;\, [\,\textit{Event}\,(e_x)\, \wedge\; \textit{At}\,(e_x,\,t_{1(x)},t_{2(x)})\,\wedge\;\textit{Event}\,(e_y)\, \\ \wedge\; \textit{At}\,(e_y,\,t_{1(y)},t_{2(y)})\,\wedge\;\bm{\textit{\textbf{Event}}\,(e)\, \wedge\;\textit{\textbf{Comprises}}\,(e_x,e_y,e)}\,]
\end{multline}
Hierarchical structures are typically reserved for specialized datasets like hierarchical action recognition benchmarks. Some string-based structures also model events hierarchically, but again, in these cases, it is typically more difficult to discern and classify them as such.

\subsection{Temporal Structure: Discussion}
The majority of datasets emphasize the timing of events: While the timing of events and their relationships to each other are often considered an afterthought in text-centric event modeling work, it is clear that it is difficult to avoid including temporal relationships in video, and so we argue that optimal methods for structuring the temporal aspect of events and formalizing them should be explored further. Along these lines, it is sensible that many datasets are compositional or hierarchical in nature. While sub-events in text can easily be summarized through higher-level events, no such analogue exists in the video medium: If an event is visible, a non-empty subset of the subevents constituting it must also be shown. Therefore, it makes sense for sub-events to be structured as hierarchical structures.
\section{Video Event Extraction Tasks}
There are multiple incipient lines of work considering video event extraction as a task, each with its own distinct task definition. We outline and analyze approaches below:

\subsection{Multimodal Event Extraction}
Chen et al. \cite{chen2021joint} argue that video events should be considered in tandem with natural language descriptions. They define Multimodal Event Extraction (MMEE) as the task of taking in a video and text sentence and classifying the event type, event trigger text span, argument text spans, and bounding boxes for each argument.

Chen et al. provide a sound video-centric extension to the text-only event extraction task. However, it does not consider compositional or hierarchical events and requires perfect event mappings to exist between text and video. Additionally, text content is required for the task, and robust event understanding necessitates the ability to identify visual events even without ground truth text documents.

\subsection{Video Semantic Role Labeling}
Sadhu et al. \cite{sadhu2021visual} build on Visual Semantic Role Labeling \cite{yatskar2016situation} through Video Semantic Role Labeling (VidSRL), the task of taking in a video and predicting a set of events as defined in VSRL (an event name and a set of related roles), the set of which constitutes a ``situation", and labeling the relationships between the events. Later, Khan et al. add bounding the visual roles to the task \cite{khan2022grounded}.

Sadhu et al.'s task of VidSRL requires no text documents as input and considers compositional events, and Khan et al.'s modifications add spatial grounding annotations. However, all events in this task are defined through 2-second intervals, which is not guaranteed to align with the temporality of the actual depicted events. Furthermore, the task only considers simple events in short clips, so VidSRL may not be optimal for long clips and higher-level events.

\subsection{Multimodal Event Hierarchy Extraction}
Ayyubi et al. \cite{ayyubi2022multimodal} build on VidSRL to construct an event hierarchy extraction task, Multimodal Event Hierarchy Extraction (MEHE): Given a text article and video, the aim is to identify all event coreferences between the two modalities and structure their hierarchical relationships through the pairwise labels ``hierarchical" and ``identical".

MEHE models both hierarchical relationships between events in video and hierarchy level mismatches between events in video and text documents. The task defines events as single shots within an edited video instead of 2s clips, but this definition is not necessarily appropriate for all data. Additionally, events are only defined through single predicates, and it also requires text documents as input.

\subsection{Discussion}
Each task definition facilitates unique aspects of robust video event understanding, and so a combination of these definitions may serve as a more universal task, such as a combination of the low-level event modeling used in MMEE, the hierarchical event modeling explored by MEHE, the ability to map visual content to natural text of both approaches, and the text document independence of VidSRL. Such a task would theoretically enable systems to identify temporally defined atomic events, consolidate the events into high-level event hierarchies, structure events at all levels through Neo-Davidsonian semantics, and map coreferences to natural language text documents, addressing all primary event modeling points considered in our event structure analysis.

As explored in our discussion of semantically complex video datasets in Section \ref{s42}, difficult benchmarks can require probabilistic inference to complete functions like making comparisons between multiple likely labels and achieving human-like social reasoning. So, for this genre of high-level understanding, mapping a video to a single hierarchy of event structures may be insufficient for empowering robust event understanding. We propose that in addition to the ideas introduced by these tasks, video event understanding models should be constructed to possess an understanding of event uncertainty for robust reasoning that better reflects human visual reasoning ability.

\section{Conclusion}
We present a comprehensive analysis of 105 event-centric video datasets, introduce formal semantic annotations to organize the implicit event structures of these datasets and assess the strengths and weaknesses of such structures. We use this survey to consider the relationship between proposed video event extraction tasks and general video event understanding research. Informed by our survey, we propose that natural language descriptions and formal event structures are both helpful for grounded event understanding, that the temporality and hierarchy of video events are critical aspects of video event modeling, and that uncertainty is a core component of modeling video events in practical and high-complexity scenarios. As video generation models improve and datasets begin to include synthetic content, we recommend that researchers producing such resources consider the visual, cultural, and presentation diversity of the generated videos.

{\small
\bibliographystyle{ieee_fullname}
\bibliography{egbib}
}

\end{document}